\documentclass{article}

\usepackage[T1]{fontenc}
\usepackage{lmodern}
\usepackage{graphicx}
\usepackage{amsmath}
\usepackage{booktabs}
\usepackage{hyperref}

\title{ReplaySAM: Zero-Shot 3D Particle-Pack Segmentation with SAM 2}
\author{Bogong Wang \quad Yulai Zhang \quad Philipp D. L\"osel\\
Andrew Kingston \quad Mark Knackstedt\\[0.5em]
\small Research School of Physics, Australian National University\\
\small Acton, ACT 2601, Australia}
\date{}

\begin{document}

\maketitle

\begin{abstract}
X-ray computed tomography (XCT) provides non-destructive three-dimensional (3D) images of particle packs, making their internal structure accessible for quantitative analysis.
Extracting particle-level measurements from these volumes first requires identifying and separating individual particles, a task that remains challenging in dense, irregular packs.
Existing workflows often require manual correction or task-specific training, while promptable models need a strategy for automatically locating individual particles and recovering their instance masks.
We present ReplaySAM, which adapts the pretrained Segment Anything Model 2 (SAM2) to three-dimensional particle-pack segmentation without additional training.
It generates point prompts from foreground geometry and fuses SAM2 predictions across orthogonal slice sequences to recover particle instances.
On XCT volumes of real quartz and iron-ore particle packs, ReplaySAM achieves higher mean instance Dice scores than expert-assisted watershed and nnInteractive under the evaluated protocols.
ReplaySAM thus provides a training-free framework for 3D particle-pack instance segmentation, with geometry-based prompting that eliminates manual prompt placement.
The source code is publicly available at \url{https://github.com/bogongwang/ReplaySAM}.
\end{abstract}


\section{Introduction}
\label{sec:intro}

Characterising individual particles in three dimensions is important in many applications ranging from materials science to mineral processing.
One important application is understanding how crushing and various separation processes affect ores during beneficiation \cite{pereira2023all, liu2018comparison, zhang2024improved}.
Particle size, shape, and mineral composition provide information for assessing processing outcomes and the liberation of valuable minerals \cite{zhang2024improved, lin20053d, reyes2018quantifying}.
X-ray computed tomography (XCT) enables non-destructive examination of particle geometry and internal structure, and a single scan can capture thousands of particles packed into a container.
Extracting particle-level measurements from these images requires particle-pack instance segmentation: identifying and assigning a distinct label to each particle throughout the reconstructed volume.
Reliable segmentation therefore connects volumetric imaging with quantitative analysis of particle populations \cite{zhang2024improved, wang2016multisize}.

Separating densely packed particles is difficult because neighbouring particles touch, while their boundaries may be weak or indistinct in the reconstructed image.
Irregular shapes, internal pores and fractures, and variation in particle size further complicate the distinction between a particle's internal structure and its external boundary \cite{zhang2024improved, sun2019watershed}.
Errors that split one particle into several instances or merge neighbouring particles directly affect the resulting particle measurements.
Manual separation and correction becomes labour-intensive and even potentially prohibitive when a tomogram contains thousands of particles, motivating methods that reduce both annotation requirements and operator intervention.

Existing approaches can be grouped into (i) classical image-processing workflows, (ii) domain-specific supervised networks, and (iii) promptable models used without task-specific training.
Marker-controlled watershed is a classical approach that separates a binary particle foreground using markers derived from its distance transform \cite{wang2016multisize,sun2019watershed,sheppard2014techniques}.
It works well when each particle produces a distinct interior marker, but irregular particles can generate multiple markers and become over-segmented, requiring manual merging or refinement \cite{zhang2024improved}.
Supervised networks instead learn appearance and boundary features from annotated data.
U-Net \cite{ronneberger2015unet}, its three-dimensional variants \cite{cicek20163dunet, chen2021transunettransformersmakestrong}, and nnU-Net \cite{isensee2021nnunet} provide established segmentation architectures and training workflows.
Their application requires suitable training labels, and changes in material appearance or acquisition conditions can motivate further annotation and adaptation.

Promptable models offer a way to use pretrained segmentation capabilities on new images without training a model for each target dataset.
SAM predicts two-dimensional masks from points, boxes, or masks, while SAM2 extends this capability to video through memory-based mask propagation \cite{kirillov2023sam,ravi2024sam2}.
For volumetric data, nnInteractive directly predicts three-dimensional masks from prompts supplied on two-dimensional views \cite{isensee2025nninteractive}.
However, promptable segmentation still requires a strategy for locating the objects to segment: manually prompting thousands of particles is burdensome, while using dense grid prompts introduces many redundant candidates.
Applying an image or video model to a tomogram also requires linking particle cross-sections across slices and resolving predictions into three-dimensional instances.
For large tomographic slices, resizing to a model's fixed input resolution can further obscure small particles.
The practical challenge is thus to combine automatic particle localisation with volumetric mask recovery while retaining the benefits of pretrained models.

In this paper, we introduce ReplaySAM, a framework that uses SAM2 to segment individual particles in three-dimensional particle packs without training.
ReplaySAM follows a three-step workflow: (i) automatic prompt generation, (ii) volumetric particle segmentation as pseudo-video segmentation in three tomographic axes, and (iii) three-dimensional mask recovery.
First, ReplaySAM uses the geometry of the particle foreground to locate candidate points within individual particles; these points serve as prompts for SAM2.
For each prompt, ReplaySAM presents the slices from each tomographic view to SAM2 as a pseudo-video sequence.
It then combines the resulting masks from the three views by majority voting to recover a three-dimensional mask for the prompted particle.
Repeating this segmentation process for all generated prompts produces the complete particle-pack instance map. 
This repeated segmentation across views and prompts gives \emph{ReplaySAM} its name.

We evaluate ReplaySAM on real XCT volumes of quartz and iron-ore and a synthetic iron-ore volume.
Under the evaluated protocols, ReplaySAM achieves the best performance on the two real volumes, exceeding the expert-assisted watershed workflow and nnInteractive.
The synthetic volume additionally provides a controlled benchmark for diagnosing ReplaySAM across different particle sizes.
On this benchmark, ReplaySAM outperforms nnInteractive, while expert-assisted watershed achieves a higher overall Dice score.
An ablation study further examines the effects of multi-view fusion, zoomed inference, and automatic prompt generation.

The paper proceeds as follows.
Section~\ref{sec:replaysam} describes the ReplaySAM methodology, including automatic prompt generation, pseudo-video segmentation, and multi-view aggregation.
Section~\ref{sec:evaluation} presents the evaluation datasets and protocols, quantitative and qualitative comparisons, ablation studies, and small-particle analysis.
Section~\ref{sec:discussion} discusses the findings, practical implications, limitations, and directions for future work.
Finally, Section~\ref{sec:conclusion} concludes the paper.

\section{ReplaySAM}
\label{sec:replaysam}

\begin{figure}[t]
    \centering
    \includegraphics[width=\textwidth]{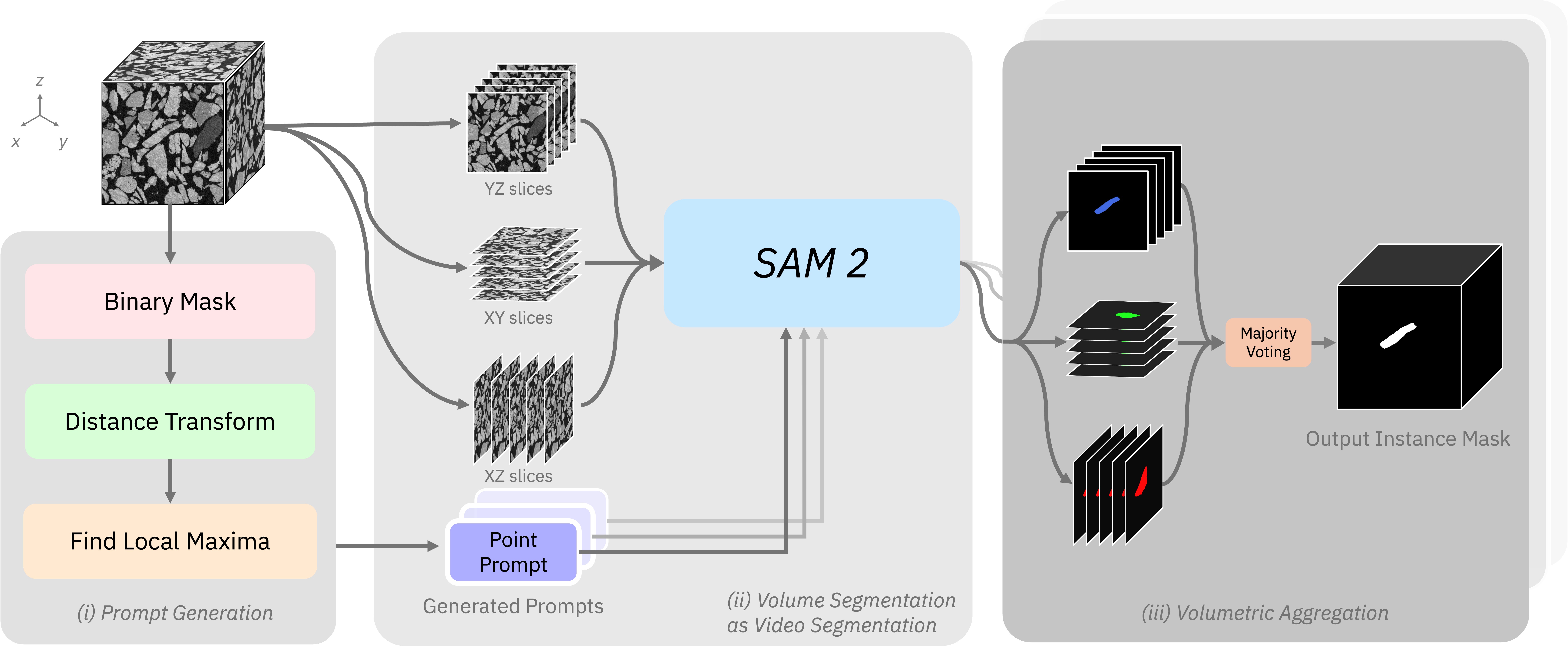}
    \caption{Overview of ReplaySAM for 3D particle-pack instance segmentation. A tomogram is binarised and transformed using a 3D Euclidean distance transform; its local maxima provide particle-interior point prompts. For each prompt, the three orthogonal slice stacks are treated as pseudo-videos and processed independently by SAM2. The resulting volumetric hypotheses are fused by voxel-wise majority voting. Fused masks are written iteratively to form the output instance map.}
    \label{fig:method-overview}
\end{figure}

Here, we present ReplaySAM, a training-free framework for 3D particle-pack instance segmentation. 
It adapts SAM2, which performs promptable 2D video segmentation, for 3D particle segmentation.
Figure~\ref{fig:method-overview} provides an overview of the complete pipeline that consists of three stages:
(i) geometry-based prompt generation: we binarise the tomogram based on a threshold and use a 3D Euclidean distance transform to identify candidate points in particle interiors as input prompts for SAM2.
(ii) single-particle slice-wise mask propagation: we convert each candidate into positive point prompts and apply SAM2 independently to the three orthogonal slice stacks, treating each stack as a pseudo-video.
(iii) multi-view fusion: we fuse the three masks by voxel-wise majority voting.
The following subsections expand on each of these stages.

\subsection{Geometry-based Prompt Generation}
\label{sec:prompt-generation}
\begin{figure}[t]
    \centering
    \includegraphics[width=.70\textwidth]{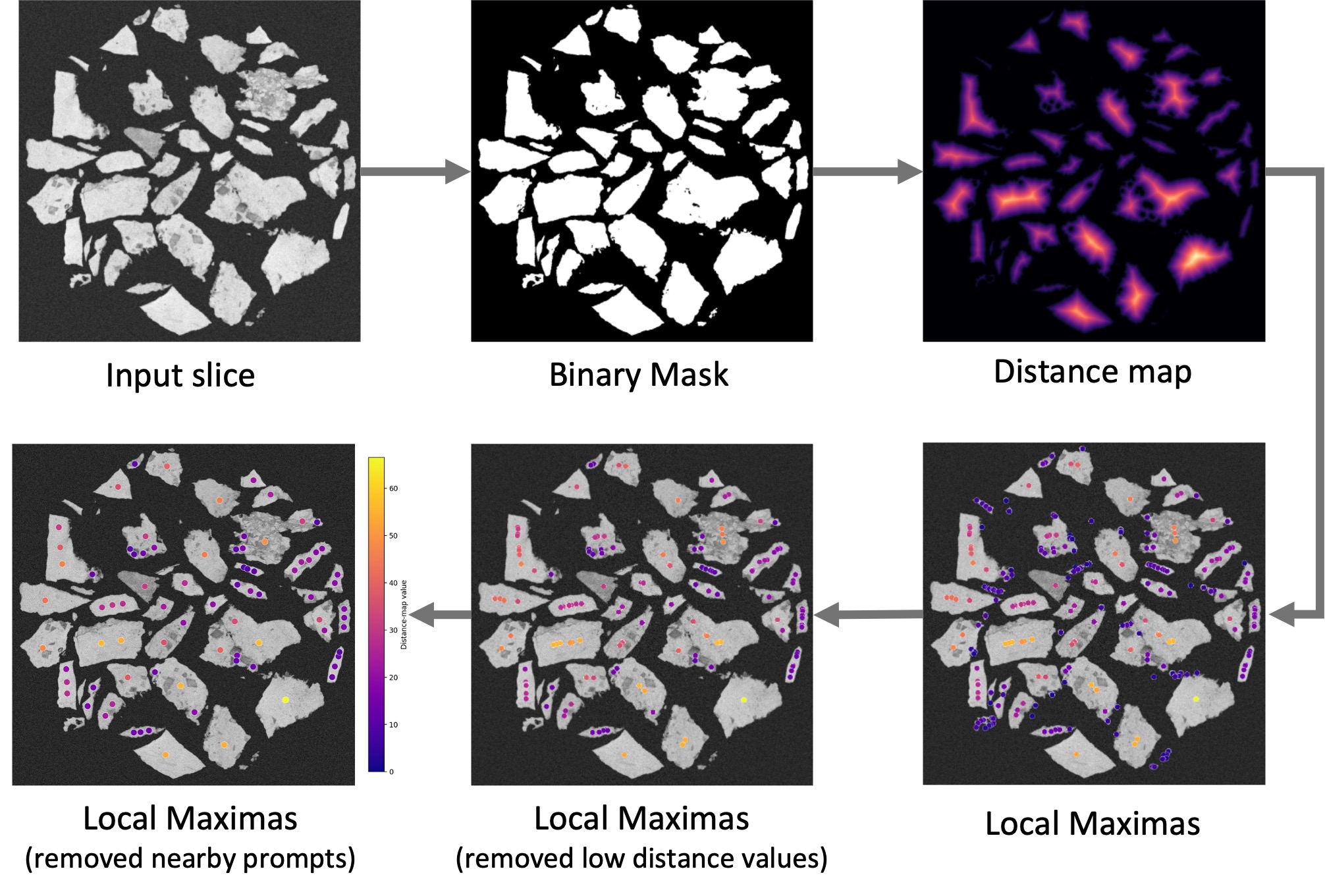}
    \caption{2D illustration of geometry-based prompt generation. 
    From an input slice, ReplaySAM derives a binary mask and its Euclidean distance map, locates local maxima, removes maxima below the distance threshold, and removes redundant nearby maxima. 
    The retained maxima serve as input point prompts for ReplaySAM. 
    The same procedure is applied to 3D tomograms.}
    \label{fig:prompt-generation-example}
\end{figure}

Similar to the marker-generation step in classical watershed segmentation, we derive point prompts from local maxima of the foreground mask's distance transform, eliminating the need for manual interaction with SAM2.
Figure~\ref{fig:prompt-generation-example} illustrates this procedure in 2D; ReplaySAM applies the same operations to the full 3D volume.

Given a grayscale tomogram, we first apply Otsu's thresholding method \cite{otsu1979threshold} to obtain a binary foreground mask and then compute its 3D Euclidean distance transform.
Each foreground voxel stores its distance to the nearest background voxel; larger values therefore correspond to locations deeper within the foreground region, which are more likely to yield reliable particle-interior prompts.

We identify candidate points within particles as local maxima of the distance map.
To reduce redundancy, we first discard candidates whose distance-map values fall below a minimum threshold, $\tau_{dist}$.
The remaining candidates are then sorted in descending order of distance-map value.
The highest-valued candidate is selected as a prompt, and all other candidates within a radius equal to its distance-map value are removed.
The candidate list is updated after each removal, and this process is repeated until no candidates remain.
This greedy procedure retains the deepest interior candidate among nearby alternatives and produces the final set of prompts for the subsequent step.

The selected prompts are processed sequentially in descending order of their distance-transform values. 
ReplaySAM maintains an occupancy volume, $S$, initially set to zero, indicating which voxels have already been assigned to accepted particle masks. 
Before segmenting a candidate point, $p_i$, we check its location in the current occupancy volume. 
If $S(p_i)=1$, the candidate is skipped because its particle has already been segmented. 
Otherwise, $p_i$ is processed by SAM2 (introduced in Section~\ref{sec:volume-segmentation}) and the resulting multi-view mask, $M_i$, is post-processed and accepted as a new instance.
This avoids redundant SAM2 inference when several distance-map maxima belong to the same particle, while processing the most reliable interior candidates first.

\subsection{Volume Segmentation as Replayed Video Segmentation}
\label{sec:volume-segmentation}
\begin{figure}[t]
    \centering
    \includegraphics[width=\textwidth]{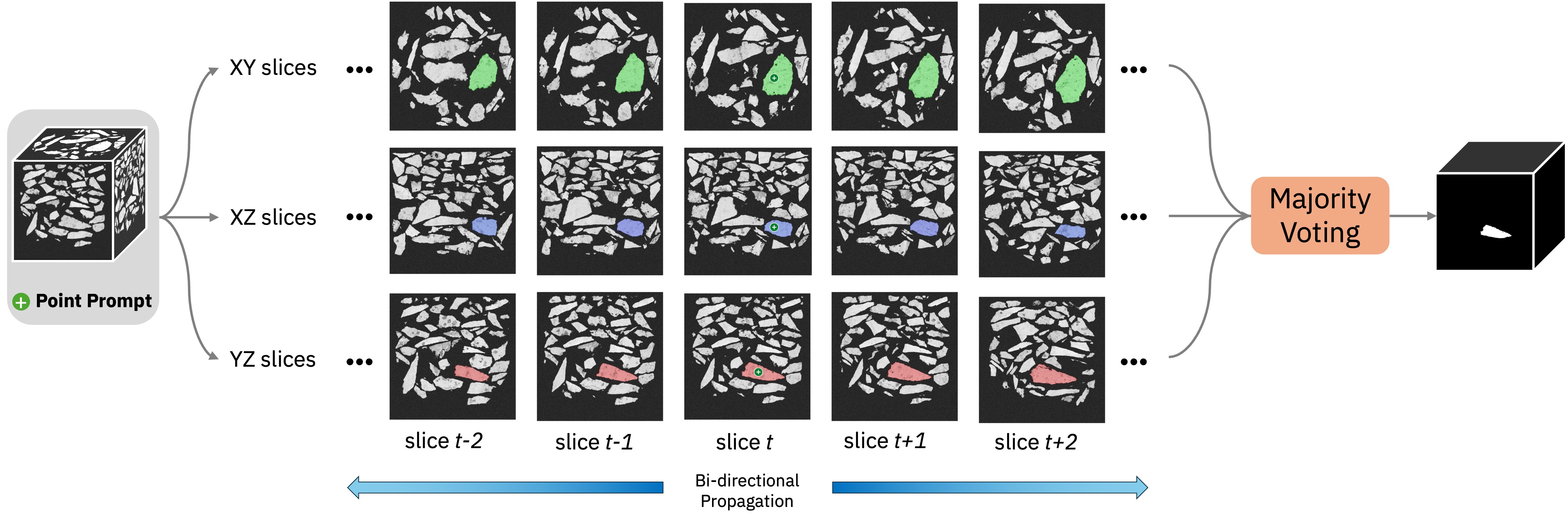}
    \caption{Reformulating single-particle volumetric segmentation as replayed video segmentation. 
    Given a 3D point prompt, ReplaySAM treats the three orthogonal slice stacks as pseudo-videos. 
    For each stack, the prompt is projected onto the corresponding 2D slice, and SAM2 propagates the particle mask bidirectionally across neighbouring slices from slice $t$. 
    The predicted slices are reassembled into three axis-specific 3D masks, which are fused by voxel-wise majority voting to produce the final particle mask.}
    \label{fig:vol-seg-as-vid-seg}
\end{figure}

For a given 3D point prompt identified in Section~\ref{sec:prompt-generation}, ReplaySAM reformulates single-particle volumetric segmentation as three video segmentation tasks, one for each tomographic axis.
Along each axis, the volume is treated as an ordered sequence of 2D slices. 
The 3D point prompt is projected onto the slice plane to obtain a 2D point prompt, which initialises SAM2's video-propagation mechanism. 
Here, slice position in volume replaces time in video, enabling SAM2 to propagate the particle mask across neighbouring cross-sections in both forward and backward directions.
Propagation terminates if the predicted mask contains fewer than \(\tau_A\) foreground pixels or if its intersection-over-union (IoU) with the mask from the last slice falls below \(\tau_{\mathrm{IoU}}\). 
Figure~\ref{fig:vol-seg-as-vid-seg} illustrates this three-axis reformulation and the subsequent fusion of the propagated masks.

Processing an entire tomographic slice at SAM2's fixed input resolution of $1024 \times 1024$ pixels can cause small particles to occupy only a few pixels when the full resolution slice is downsampled.
ReplaySAM therefore uses coarse-to-fine zoomed inference: a coarse pass first estimates the particle's approximate location and boundary, after which a fine pass segments the localised region at a higher effective resolution.
The refined mask then initialises SAM2's bidirectional propagation through the remaining slices.

Using multiple orientations reduces reliance on the appearance of the particle in any single cross-sectional view.
A particle boundary that is ambiguous or difficult to propagate in one orientation may be more clearly represented in another, making the three predictions complementary.

Bidirectional propagation is necessary because the automatically generated candidate point usually lies near the particle centre; propagating in both slice directions is therefore required to recover the particle's full extent.

ReplaySAM therefore applies SAM2's pretrained video-segmentation capability to volumetric particle segmentation without modifying the model architecture or training it on particle-specific masks.

\subsection{Multi-view Volumetric Aggregation}
Multi-view fusion is an established 2.5D strategy for integrating predictions from orthogonal slice stacks, with majority voting providing a conventional rule for combining the views \cite{zhang2022bridging,li2020modeldriven}.
Processing the three orthogonal slice sequences produces three volumetric masks for the prompted particle.
We aggregate the three segmented masks by voxel-wise majority voting.

Let $M_1(\mathbf{x}), M_2(\mathbf{x}), M_3(\mathbf{x}) \in \{0,1\}$ 
denote the three axis-specific predictions at voxel $\mathbf{x}$, where 
$1$ denotes foreground. We obtain the fused mask by retaining voxels 
classified as foreground in at least two views:
\begin{equation}
    M_{\mathrm{fused}} =
    \left\{
    \mathbf{x}\;\middle|\;
    \sum_{i=1}^{3} M_i(\mathbf{x}) \geq 2
    \right\}.
    \label{eq:majority-fusion}
\end{equation}

This aggregation limits the influence of one unreliable orientation by removing regions that appear in only one view while preserving regions supported by at least two views.
The result is one fused mask for the candidate particle.

Applying this single particle segmentation procedure to all automatically generated prompts, in descending order of their distance-map values, produces the complete particle-pack instance map.
Alternatively, automatic prompt generation can be bypassed, allowing users to interactively segment a selected particle with a point prompt.

\section{Evaluation}
\label{sec:evaluation}

We evaluate ReplaySAM as a zero-shot pipeline for particle instance segmentation in several different three-dimensional tomograms for particle packs.
The evaluation addresses four questions: 
(i) how automatic prompt generation affects the accuracy and efficiency of the pipeline;
(ii) how ReplaySAM compares with an expert-assisted watershed workflow and nnInteractive across the evaluated materials; 
(iii) how multi-view aggregation and zoomed inference contribute to segmentation performance; and
(iv) what the qualitative examples reveal about ReplaySAM's segmentation behaviour.

\subsection{Evaluation Method}
\label{sec:experimental-setup}

To address the above questions, we first describe the evaluation methods and settings, the datasets, and the evaluation metric.
This is useful for comparing the methods consistently, interpreting their performance on the selected data, and reproducing the experiments.

\subsubsection{Evaluation Settings}
\label{sec:evaluation-settings}

We compare three approaches spanning an established expert-assisted workflow and modern promptable segmentation: advanced watershed, nnInteractive, and the proposed ReplaySAM. 
Advanced watershed is included because watershed-based workflows remain the most common approach to particle-pack segmentation \cite{zhang2024improved, wang2016multisize, sun2019watershed, barron2025techniqueshighspeedimagingxray}. 
nnInteractive is included as a state-of-the-art interactive method that performs promptable segmentation natively in three dimensions \cite{isensee2025nninteractive}.

In this study, \emph{advanced watershed} refers to marker-controlled watershed followed by manual post-processing \cite{zhang2024improved}.
The automatic stage of this watershed method first binarises the tomogram and computes the three-dimensional Euclidean distance transform of the foreground.
Local maxima of the distance map serve as particle-interior markers for the watershed separation of contacting particles.
Because marker generation can leave adjacent particles merged or split an irregular particle into multiple regions, the resulting instance map is manually inspected, and over-segmented regions with a large shared interfacial area are merged \cite{zhang2024improved}.
This comparator therefore represents the established expert-assisted workflow used in practice rather than an unattended watershed implementation.

nnInteractive is evaluated with its default settings as the native-3D interactive comparator.
Given a prompt specified within the volume, it directly predicts a three-dimensional mask of the target particle.
To ensure a consistent prompting protocol, nnInteractive receives the same automatically generated particle-interior point used to initialise the corresponding ReplaySAM prediction.

All ReplaySAM experiments use the same components and parameter settings described below.
It generates particle-interior prompts automatically with the minimum Euclidean distance threshold \((\tau_{dist})\) set to 5 voxels. 
For each prompt, SAM2 propagates an initialised mask through three orthogonal slice sequences using the proposed zooming procedure; the resulting predictions are then fused by majority voting.
The pretrained \texttt{SAM2.1 Hiera-Large} model is used as the backbone throughout because it achieves the strongest video-segmentation performance among the evaluated SAM 2 model variants on the benchmarks reported in the original SAM 2 study \cite{ravi2024sam2}.
The minimum mask-area threshold is set to \(\tau_A=10\) pixels, and the propagation termination threshold is set to \(\tau_{\mathrm{IoU}}=0.7\).
For evaluation, only segmented instances containing more than 100 voxels are retained.

These thresholds were chosen empirically as heuristic operating values; no formal hyperparameter optimisation was performed.
Once selected, the same values were held fixed across all datasets and experiments.
The setting \(\tau_{dist}=5\) voxels suppresses shallow distance-transform maxima near uncertain foreground boundaries without imposing a larger minimum interior-distance requirement that would exclude more small-particle prompts.
The setting \(\tau_A=10\) pixels terminates propagation when the predicted cross-section becomes too small to provide reliable evidence, limiting spurious masks and propagation beyond the particle extent.
The setting \(\tau_{\mathrm{IoU}}=0.7\) permits moderate changes in cross-sectional shape between adjacent slices while terminating propagation after an abrupt reduction in overlap, which is likely to indicate mask drift or propagation failure.
The 100-voxel minimum removes tiny predicted fragments that are more likely to be segmentation artefacts than meaningful particle instances.
Because these thresholds are expressed in pixels or voxels, they may require adjustment for datasets with substantially different spatial resolutions or particle-size distributions.

Runtime evaluation measures the time taken by each method to complete the segmentation task.
For ReplaySAM and nnInteractive, we measure end-to-end wall-clock time on a single workstation equipped with an Intel Core i5-13600 CPU, 64\,GB of RAM, and a single RTX 4090 GPU.
This includes the time for prompt generation, SAM2 inference or nnInteractive inference, and post-processing.
For advanced watershed, we report only the elapsed computation time of the processing steps executed on a high-performance computing cluster.
These measurements exclude time spent selecting parameters, inspecting segmentation quality, and rerunning a segmentation following an unsatisfactory result.
This computational environment was necessary because the workflow's memory requirements exceeded the practical capacity of the evaluation workstation and several processing steps relied on parallel CPU execution. 
CPU and memory allocations in advanced watershed were adjusted to the requirements of individual workflow steps and therefore cannot be represented by a single hardware specification.

\subsubsection{Datasets}
\begin{figure}[t]
    \centering
    \includegraphics[width=\textwidth]{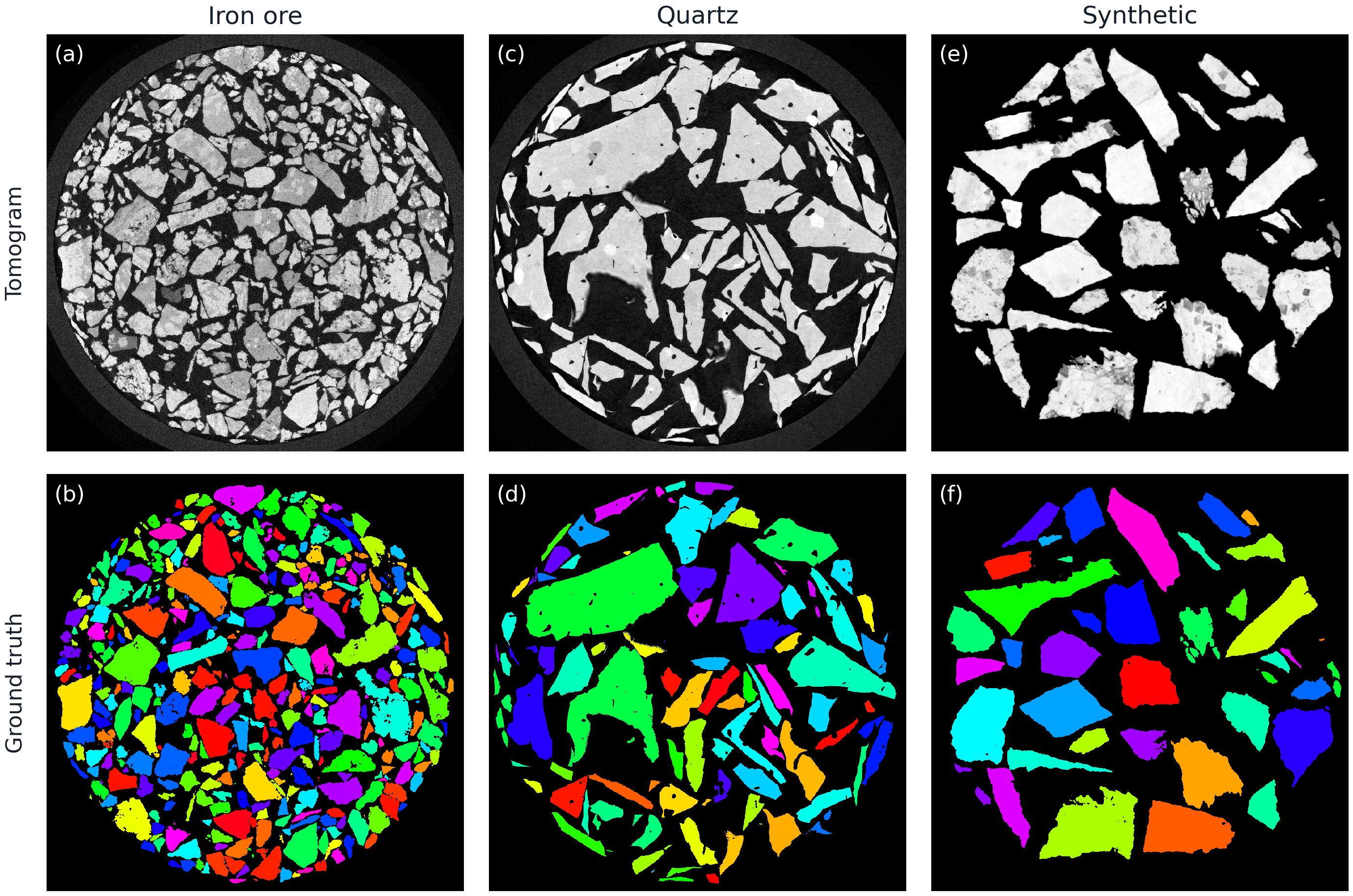}
    \caption{Example tomogram slices (top row) and corresponding ground-truth instance masks (bottom row) for the iron-ore, quartz, and synthetic datasets.}
    \label{fig:datasets-preview}
\end{figure}

\begin{table}[t]
    \centering
    \small
    \begin{tabular}{llll}
        \toprule
        Dataset    & Material             & Volume size                    & Instances \\
        \midrule
        iron-ore   & Iron Ore             & $4004 \times 2802 \times 2802$ & 7136 \\
        quartz     & Quartz               & $5457 \times 2424 \times 2424$ & 2880 \\
        synthetic  & Synthetic (Iron Ore) & $1024 \times 1024 \times 1024$ & 500 \\
        \bottomrule
    \end{tabular}
    \caption{Evaluation volumes. 
    Volume size denotes the dimensions of each 3D volume in voxels. 
    ``Instances'' denotes the number of reference particle instances (i.e., particle masks) included in each volume.}
    \label{tab:datasets}
\end{table}

\begin{table}[t]
    \centering
    \small
    \begin{tabular}{lcc}
        \toprule
        Size class & Actual volume $V$ ($\mathrm{voxels}^3$) & Equivalent-sphere radius $r_{\mathrm{eq}}$ (voxels) \\
        \midrule
        Small  & $<50{,}000$              & $<23$ \\
        Medium & $50{,}000$--$500{,}000$ & $23$--$49$ \\
        Large  & $>500{,}000$             & $>49$ \\
        \bottomrule
    \end{tabular}
    \caption{
    Particle-size classes for the synthetic-volume evaluation. 
    For scale interpretation, we treat each object as a sphere with the same volume, $r_{\mathrm{eq}}=\left(\frac{3V}{4\pi}\right)^{1/3}$. 
    This approximation is illustrative only; class assignment uses each object's actual voxel volume.}
    \label{tab:particle-size-classes}
\end{table}

Our evaluation uses XCT volumes of quartz and iron-ore particle packs, together with a synthetic particle-pack volume constructed from real iron-ore particles.

The experimental packings of quartz and iron-ore particles were produced using two different crushing processes. 
Iron ore particles were produced from a packing of parent particles that were crushed under quasi-static (typically over 2 hours) oedoemetric loading conditions \cite{francois2022versatile, kandula2024situ}. 
Quartz particles were produced using high-energy impact crushing on a single parent particle; this fragmentation process occurs over a millisecond time scale \cite{barron2025techniqueshighspeedimagingxray}. 
These different crushing processes (quasi-static and dynamic) produced fragmented particles with a broad range of sizes and morphologies \cite{barron2025techniqueshighspeedimagingxray}.
In both cases, the particles were prepared following the procedure described in \cite{barron2025techniqueshighspeedimagingxray}, packed into containers, and scanned using CT.
The resulting high-resolution scans constitute the datasets used in this study.

In addition, the distinct characteristics of these materials pose different challenges for segmentation.
Iron-ore particles have complex internal structures, including porosity and fractures, that can make their outer boundaries difficult to distinguish from internal features.
Quartz particles from a crushed sintered quartz bead are commonly elongated and plate-like, allowing them to pack tightly and form extended contacts for which the boundary between adjacent particles is difficult to identify.
Both material groups contain a range of particle sizes, challenging ReplaySAM's ability to perform robust segmentation across varying size distributions.
Table~\ref{tab:datasets} summarises their material types, image dimensions, and numbers of annotated instances.

The synthetic iron-ore volume and its instance segmentation were generated using a physics-based simulation \cite{wang2025syntheticparticlepack}. This dataset provides a controlled benchmark with a fixed particle population, voxel resolution, and exact ground-truth masks.
The 500 evaluated particles were stratified by their voxel volumes, with approximately one third assigned to each size class in Table~\ref{tab:particle-size-classes}. 
To make the scale of these 3D volumes easier to interpret, the table also reports the radius of a sphere with an equivalent volume, providing an intuitive indication of particle size.
This spherical representation is used only for illustration; particles are assigned to classes according to their actual 3D volumes, irrespective of shape. 
These volume-based classes enable a targeted analysis of small-particle recovery.

Creating comparably reliable reference labels for the real tomograms is more difficult.
Because each real tomogram contains thousands of particles, exhaustive manual segmentation is impractical.
We therefore use the segmentations generated by the self-validated learning (SVL) method \cite{losel2025selfvalidatedlearningparticleseparation} implemented in Biomedisa \cite{losel2020introducing} as a surrogate ground truth.
The resulting label volumes are manually inspected before being used for evaluation.

\subsubsection{Metric}

We use mean instance Dice as the quantitative metric because it has become standard in volumetric segmentation tasks \cite{isensee2021nnunet, isensee2025nninteractive, hatamizadeh2021unetrtransformers3dmedical}.
Each ground-truth particle is matched to the predicted particle that overlaps it most closely, as measured by the Dice score. 
Each prediction can be matched only once, so a single prediction covering several particles cannot be counted as a successful match for all of them.

For a matched prediction--ground-truth pair $(P_i,G_i)$, the Dice score is
\begin{equation}
    \operatorname{Dice}(P_i,G_i)
    = \frac{2\lvert P_i \cap G_i\rvert}{\lvert P_i\rvert + \lvert G_i\rvert}.
    \label{eq:instance-dice}
\end{equation}
An unmatched ground-truth instance receives a Dice score of zero, and the dataset score is the arithmetic mean over all ground-truth instances.

\subsection{Comparison of Prompt-Generation Strategies}
\label{sec:prompt-generation-comparison}

\begin{table}[t]
    \centering
    \small
    \begin{tabular}{lccc}
        \toprule
        Prompt strategy         & Generated prompts & Dice         & Total time (min) \\
        \midrule
        Uniform grid            & $1{,}000{,}000$ & 0.784          & 1501 \\
        Geometry-based (ours)   & $27{,}620$      & \textbf{0.803} & \textbf{131} \\
        \bottomrule
    \end{tabular}
    \caption{Comparison of prompt-generation strategies on the synthetic iron-ore volume.
    The uniform baseline uses a $100^3$ grid, yielding $10^6$ input prompts. 
    Total time is the wall-clock time in minutes, including prompt generation and subsequent SAM2 inference.
    The best Dice score and shortest total time are bold.
    }
    \label{tab:prompt-generation-comparison}
\end{table}

We test the proposed geometry-based prompt generator against uniform-grid sampling, a strategy used by SAM for automatic point-prompt generation \cite{kirillov2023sam}.
This comparison evaluates segmentation accuracy and computational efficiency on the synthetic iron-ore volume, which contains 500 reference particle instances and has dimensions of $1024 \times 1024 \times 1024$ voxels (Table~\ref{tab:datasets}).
The uniform-grid baseline uses a $100^3$ grid, yielding $10^6$ input prompts, whereas the proposed geometry-based generator uses the default configuration described in Section~\ref{sec:evaluation-settings}. 
Both strategies were evaluated using the same hardware configuration and post-processing settings specified in that section.
Table~\ref{tab:prompt-generation-comparison} reports mean instance Dice, the number of generated prompts, and end-to-end runtime.

The geometry-based generator improves Dice from 0.784 to 0.803 while reducing the number of generated prompts from $1{,}000{,}000$ to $27{,}620$.
Its end-to-end runtime is 131 min, compared with 1501 min for the uniform-grid baseline.

\subsection{Main Quantitative Comparison}
\label{sec:main-results}

This experiment assesses whether ReplaySAM can achieve competitive 3D particle-instance segmentation accuracy without task-specific training or manual prompt placement.
We compare ReplaySAM with advanced watershed and nnInteractive, representing an established expert-assisted workflow and a native-3D promptable model, respectively, across the real quartz and iron-ore tomograms and the controlled synthetic volume.
As detailed in Section~\ref{sec:evaluation-settings}, nnInteractive and ReplaySAM were executed on the single evaluation workstation, whereas advanced watershed was executed on an HPC cluster with CPU and memory resources varying among workflow steps.

\begin{table}[t]
    \centering
    \begin{tabular}{lcccc}
        \toprule
        Method            & iron-ore       & quartz          & synthetic \\
        \midrule          
        Advanced watershed& 0.925          & 0.932           & \textbf{0.971} \\
        nnInteractive     & 0.726          & 0.819           & 0.701 \\
        ReplaySAM         & \textbf{0.933} & \textbf{0.942}  & 0.803 \\
        \bottomrule
    \end{tabular}
    \caption{Main comparison on the three evaluation volumes. Entries are mean instance Dice (higher is better); the best score in each column is bold.}
    \label{tab:main-results}
\end{table}

\begin{table}[t]
    \centering
    \small
    \begin{tabular}{lccc}
        \toprule
        Method             & iron-ore     & quartz      & synthetic \\
        \midrule
        Advanced watershed\textsuperscript{*} & \textbf{106} & \textbf{59} & \textbf{4} \\
        nnInteractive      & 1488         & 647         & 40   \\
        ReplaySAM          & 2477         & 1466        & 131   \\
        \bottomrule
    \end{tabular}
    \caption{Runtime comparison on the three evaluation volumes in minutes. The nnInteractive and ReplaySAM entries are end-to-end wall-clock runtimes measured on a single workstation. \textsuperscript{*}The advanced-watershed entries report processing time only on a high-performance computing cluster, with CPU and memory resources varying among workflow substeps. They exclude time spent selecting parameters, inspecting segmentation quality, and rerunning a segmentation after an unsatisfactory result.}
    \label{tab:runtime-comparison}
\end{table}

Table~\ref{tab:main-results} compares the three segmentation methods across the three evaluation volumes. 
ReplaySAM achieves the highest mean instance Dice on the quartz and iron-ore volumes, reaching 0.942 and 0.933, respectively. 
These scores exceed nnInteractive by 0.123 on quartz and 0.207 on iron-ore, and exceed the expert-assisted watershed workflow by 0.010 and 0.008, respectively.
Thus, on the two real XCT volumes, ReplaySAM provides the strongest mask-overlap accuracy among the evaluated methods without task-specific training.

On the synthetic volume, ReplaySAM improves over nnInteractive (0.803 versus 0.701) but remains below advanced watershed (0.971).
Section~\ref{sec:small-particles} further discusses this result by particle size and evaluates zoomed inference as a mitigation for small-particle errors.

Table~\ref{tab:runtime-comparison} reports the runtime of the three methods under their respective computational settings.
ReplaySAM requires 2477, 1466, and 131 min for the iron-ore, quartz, and synthetic volumes, respectively, whereas nnInteractive requires 1488, 647, and 40 min.
ReplaySAM is therefore slower than nnInteractive on all three volumes, but achieves higher mean instance Dice under the same evaluation protocol.
The reported runtimes for the advanced watershed workflow include only elapsed processing time, excluding time spent selecting parameters, inspecting segmentation quality, and rerunning segmentation after an unsatisfactory result. 
Some steps, including converging active contour segmentation \cite{sheppard2004techniques,schluter2014image} and distance map watershed, commonly require more than 1 TB of memory for volumes the size of the quartz dataset (reported in Table~\ref{tab:datasets}), making local execution impractical.
The workflow was therefore run on an HPC cluster, with resources varying among substeps. Consequently, its reported runtime is not directly comparable on a hardware-normalised basis with the single-workstation runtimes of nnInteractive and ReplaySAM.

\subsection{Ablation of ReplaySAM Components}
\label{sec:ablations}

\begin{table}[t]
    \centering
    \begin{tabular}{lccccc}
        \toprule
        Variant                  & \# Axes      & Vote $k$       & Zoom           & Dice \\
        \midrule
        Best single axis         & \underline{1} & --            & yes            & 0.702 \\
        Intersection fusion      & 3             & \underline{3} & yes            & 0.693 \\
        Union fusion             & 3             & \underline{1} & yes            & 0.782 \\
        No zoom                  & 3             & 2             & \underline{no} & 0.727 \\
        ReplaySAM (default)      & 3             & 2             & yes            & \textbf{0.803} \\
        \bottomrule
    \end{tabular}
    \caption{Ablation on the synthetic volume. 
    \# Axes denotes the number of orthogonal axes used for inference. 
    Vote $k$ denotes the minimum number of axis-wise masks that must contain a voxel for it to be retained in the fused mask.
    The default ReplaySAM configuration uses three axes, majority voting ($k=2$) and zoomed inference.
    Entries are mean instance Dice (higher is better).}
    \label{tab:ablation}
\end{table}

To determine which components contribute most to successful segmentation, we ablate multi-view aggregation and zoomed inference on the synthetic volume, using the complete ReplaySAM configuration as the reference.
We use a synthetic volume for these comparisons because it provides exact instance labels and a controlled particle population. 
This lets us compare the settings under the same evaluation conditions.

The \emph{single-axis} setting tests whether aggregating predictions from multiple views improves performance over inference along one tomographic axis. 
We run three independent inferences, one along each axis, and report the highest mean Dice score. 
This provides a strong single-axis baseline for assessing the benefit of multi-view aggregation.

The \emph{union} and \emph{intersection} settings test how the fusion rule affects segmentation. 
We apply both rules to the same three axis-wise predictions. 
Union includes voxels predicted by at least one view, whereas intersection includes only voxels predicted by all three. 
Comparing these rules shows how disagreement between views influences the final result.

The \emph{no-zoom} setting tests the contribution of zoomed inference, particularly for small particles that are poorly resolved at full-resolution. 
It retains the prompts and fusion rule of the full configuration while disabling zoomed inference, allowing us to assess the effect of zooming.

All components not explicitly varied retain the configuration described in Section~\ref{sec:experimental-setup}, including the SAM2 model variant, prompt-generation procedure, propagation thresholds, post-processing, and evaluation protocol.

As shown in Table~\ref{tab:ablation}, the complete ReplaySAM configuration achieves a mean Dice of 0.803.
The best single-axis result is 0.702 ($-0.101$), while intersection and union fusion achieve 0.693 and 0.782, respectively.
Removing zoomed inference lowers Dice to 0.727 ($-0.076$).
Thus, the default configuration significantly outperforms every evaluated ablation on the synthetic volume.

\subsection{Resolution and Small-Particle Analysis}
\label{sec:small-particles}

\begin{table}[t]
    \centering
    \small
    \begin{tabular}{lcc}
        \toprule
        Size   & w/o zoom & w/ zoom \\
        \midrule
        Small  & 0.218  & 0.463 \\
        Medium & 0.973  & 0.975 \\
        Large  & 0.982  & 0.976 \\
        \bottomrule
    \end{tabular}
    \caption{Mean instance Dice for ReplaySAM by particle-size class, comparing the same configuration without and with zoomed inference on the synthetic dataset.}
    \label{tab:size-results}
\end{table}

Zoomed inference is intended to preserve appearance for particles that become too small to resolve reliably after a high-resolution tomographic slice is downsampled to SAM2's input resolution.
We therefore split ground-truth instances using the small, medium, and large volume classes defined in Table~\ref{tab:particle-size-classes}.
Within each class, we compare ReplaySAM without zoom and ReplaySAM with zoom using mean Dice.

From Table~\ref{tab:size-results}, we observe that zooming increases the mean Dice of small particles from 0.218 to 0.463, an absolute gain of 0.245.
In contrast, it has a negligible positive effect for medium particles (0.973 to 0.975; $+0.002$) and slightly reduces Dice for large particles (0.982 to 0.976; $-0.006$).
The largest absolute change is therefore observed for the small-particle subset.

\subsection{Qualitative Results}
\label{sec:qualitative-results}

\begin{figure}[t]
    \centering
    \includegraphics[width=\textwidth]{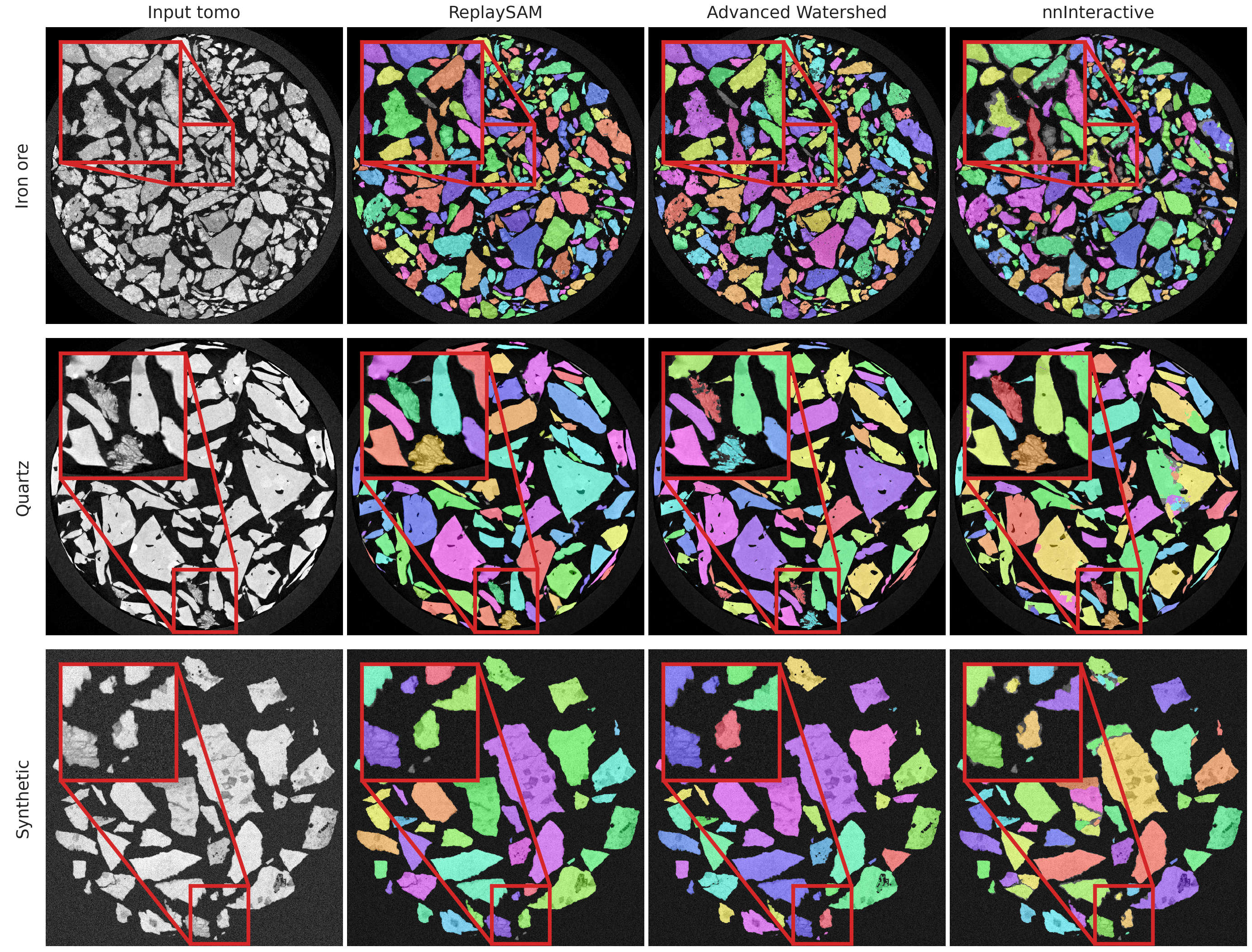}
    \caption{Qualitative comparison on representative slices from the iron-ore, quartz, and synthetic volumes (top to bottom). Columns show the input tomogram and the corresponding instance maps from ReplaySAM, advanced watershed, and nnInteractive. Each colour denotes a predicted instance within a method and is not intended to establish instance correspondence across methods.}
    \label{fig:qualitative}
\end{figure}

Figure~\ref{fig:qualitative} gives a qualitative comparison across the three evaluation materials.
Each row shows a representative tomogram slice followed by the instance maps produced by ReplaySAM, advanced watershed, and nnInteractive.
The coloured regions denote separate predicted particle instances; colours are used only for visualisation and do not encode correspondence between methods.

For iron ore, ReplaySAM captures the overall particle arrangement in densely packed regions while preserving the irregular shapes visible in the input tomogram. 
However, some small particles in the zoomed region are only partially segmented. 
Their ambiguous appearance across orthogonal slice planes may lead to inconsistent predictions, causing majority voting to retain only the portions supported by multiple views. 
Advanced watershed also misses a small, thin particle in this region, potentially because its limited thickness produces low distance-transform values that are insufficient to generate a seed under the chosen seed-selection settings. 
nnInteractive leaves the largest number of visible particles unsegmented among the three methods in this example.

For quartz, ReplaySAM segments most particles but misses a small particle in the zoomed region. 
In the advanced watershed result, the red and cyan instances cover only parts of the corresponding particles. 
These incomplete segmentations illustrate the sensitivity of watershed to the initial binary foreground mask: an unsuitable intensity threshold can exclude portions of a particle before instance separation, restricting the segmentation completeness of the particle instance. 
nnInteractive oversegments some large particles near the middle-right of the slice, splitting individual particles into multiple predicted instances.

For the synthetic material, advanced watershed produces the most complete segmentation in the displayed example, consistent with the quantitative results in Table~\ref{tab:main-results}. 
ReplaySAM captures the larger particles but misses some small particles, as highlighted in the zoomed region. 
nnInteractive recovers some of these small particles but also splits larger particles into multiple instances, including those near the centre of the slice. 
ReplaySAM's small-particle omissions are consistent with the size-stratified results in Table~\ref{tab:size-results}, where zoomed inference improves small-particle Dice but does not close the performance gap relative to medium and large particles.

\section{Discussion}
\label{sec:discussion}

This section interprets the evaluation results to clarify ReplaySAM's strengths, the factors influencing its performance, and its suitability for practical XCT particle analysis.
We discuss performance across the real and synthetic particle pack volumes, the contributions of the main design choices, and the implications for practical use, before outlining the remaining limitations and priorities for future work.

\subsection{Performance Across Evaluation Volumes}
\label{sec:performance-discussion}

ReplaySAM's comparative performance depends on the characteristics of the evaluated volume.
It achieves the highest Dice score on both real XCT volumes among the evaluated methods, while advanced watershed remains clearly stronger on the synthetic volume.

However, results on the synthetic test dataset suggest that ReplaySAM does not consistently outperform an expert-assisted watershed workflow.
Table~\ref{tab:size-results} shows that ReplaySAM's performance varies substantially with particle size, with small particles remaining the greatest challenge.
The low Dice scores for small particles are the main reason for ReplaySAM's lower mean instance Dice on the synthetic volume in Table~\ref{tab:main-results}, as each ground-truth instance contributes equally to the reported mean.
In contrast, advanced watershed separates instances from a binary foreground mask; once a reliable binary mask has been established, this separation process may be comparatively less sensitive to particle size.

While this size dependence highlights a limitation of ReplaySAM, the strong performance of advanced watershed may also reflect characteristics of the synthetic benchmark that favour its segmentation workflow.
Two properties of the synthetic dataset may contribute to this difference.
First, although the synthesis workflow produces visually realistic particle packs, it does not yet reproduce partial-volume artefacts at particle borders.
Such artefacts make binary mask construction and watershed separation more difficult in real XCT data. 
In particular, partial-volume effects cause voxels near particle interfaces to contain a mixture of particle and background intensities, reducing boundary contrast and making adjacent particles appear connected. 
This obscures the inter-particle boundaries needed for reliable thresholding and watershed separation. 
Their absence in the synthetic volume therefore likely favours watershed performance as it is easier to produce an accurate binary mask, giving watershed a better starting point for separating particles.
Second, the particle instances used to construct the synthetic benchmark were themselves obtained using a watershed-based procedure \cite{wang2025syntheticparticlepack}.
This shared algorithmic structure may introduce method-aligned bias through particle geometries or boundaries inherited from the generation workflow.
The synthetic benchmark is therefore useful for controlled comparison, but its watershed advantage should not be interpreted as fully independent evidence of generalisation.

\subsection{Effects of ReplaySAM Design Choices}
\label{sec:design-discussion}

The ablation results indicate that the three orthogonal views provide complementary evidence and that majority voting suppresses some single-view errors and thereby improves Dice.
A single viewing orientation can contain view-dependent ambiguity or incomplete boundary information that the other orientations resolve.
Intersection fusion is too restrictive because a region missed in even one view is removed from the fused mask.
Union fusion instead retains errors from any view; for example, if one view merges the target with a neighbouring particle, the union preserves the erroneous extra region.
Majority voting suppresses these single-view errors without discarding regions absent from only one view.

The size-stratified results support the motivation for zoomed inference while showing that its benefits vary with particle size. 
Zooming is most effective for small particles because it increases their effective resolution. 
When a high-resolution tomographic slice is downsampled to SAM2's input resolution, small particles occupy few pixels and may lose important boundary details.
Enlarging the corresponding regions restores some of this information and improves segmentation. 
By contrast, medium and large particles are already represented at sufficient resolution, so zooming provides little additional information. 
It may also introduce errors when an inaccurate coarse mask produces a crop that excludes part of the particle. 
Because these particles have less to gain from increased resolution, such cropping errors can offset the limited benefit of zooming. 
The overall improvement from zoomed inference is therefore driven primarily by the small-particle subset.

Compared with uniform grid sampling, the geometry-based prompt generator improves both segmentation quality and computational efficiency.
It places prompts inside particles, typically near their centres. 
In contrast, uniform grid sampling may place prompts in the background, miss small particles, or place prompts near particle boundaries. 
These cases can produce missed or inaccurate instances.
The geometry-based generator also reduces the candidate set from \(10^6\) grid points to 27,620 prompts. 
This \(36\times\) reduction in the number of prompts corresponds to an \(11\times\) speedup rather than a proportional reduction in runtime.
The generated prompts are only candidates for inference: as described in Section~\ref{sec:prompt-generation}, a candidate is skipped when it falls within a region that has already been segmented.
Because this online suppression is applied to both prompt-generation strategies, the number of generated prompts does not equal the number of SAM2 inference runs.
Consequently, prompt count and runtime are not expected to scale linearly.

\subsection{Practical Implications}
\label{sec:practical-implications}
Taken together, the results position ReplaySAM as a training-free approach to particle-instance segmentation in XCT volumes.
It is particularly relevant when task-specific annotations are unavailable or when changes in material or acquisition settings would otherwise require model retraining.

The runtime results in Table~\ref{tab:runtime-comparison} should be interpreted together with each method's interaction and computational-resource requirements.
Unlike the advanced watershed workflow, ReplaySAM does not require manual inspection or correction.
Advanced watershed was executed on an HPC cluster, with CPU and memory resources varying among workflow steps, whereas ReplaySAM was executed on the evaluation workstation.
Although ReplaySAM is slower than nnInteractive, it achieves higher mean instance Dice on all three evaluated datasets under the evaluated protocol.
Runtime alone therefore does not capture the trade-off among segmentation accuracy, operator effort, and computational-resource requirements.

ReplaySAM can be deployed in either automatic or semi-automatic workflows.
In automatic mode, it provides an initial instance map for subsequent particle-size, shape, and packing analyses, or for visual inspection and manual refinement.
This mode can support large XCT volumes for which exhaustive manual delineation is impractical.
In semi-automatic mode, automatic prompt generation can be bypassed and a user-supplied point prompt can be used to extract or correct a selected particle.
This mode is therefore suitable for targeted particle inspection and segmentation correction.

The size-stratified results also provide practical guidance for selecting an inference strategy.
Zoomed inference can be omitted when particles remain well resolved after the complete tomographic slice is resized to SAM2's input resolution.
For volumes containing particles of similar size, a fixed prompt-centred window matched to the average particle scale can replace adaptive coarse-to-fine zooming.
Current zooming strategies are most useful when particle sizes vary substantially, particularly when some particles become poorly resolved during full-slice inference.

\subsection{Limitations and Future Work}
\label{sec:limitations-future-work}

The main limitation is the segmentation of small particles. 
For sufficiently small particles, the object still occupies only a few pixels after each slice is zoomed, leaving limited boundary evidence for both initial prediction and slice-wise propagation in every orientation. 
A potential solution is to separate smaller particles physically by sieving and scan each size fraction independently at higher resolution, thereby improving their effective resolution and segmentation accuracy.

ReplaySAM also depends on the quality of the foreground binarisation and the geometry-based prompts derived from it. 
Low contrast, partial-volume artefacts, or internal particle structure can distort the distance transform and cause a particle to receive no suitable prompt or a suboptimal one. 
A possible solution is to replace the current foreground-binarisation procedure with a more sophisticated binary segmentation model, such as a U-Net, to produce higher-quality binary masks.
This could improve the reliability of the distance transform and thereby produce more accurate geometry-based prompts.

The current implementation also processes prompts sequentially.
This ordering avoids redundant predictions and conflicting instance assignments because each accepted mask is written to the output before later candidates are considered, but it makes the total runtime grow linearly with the number of retained prompts and the size of the volume. 
The axis-wise predictions for a prompt, and predictions for candidates that do not overlap, offer opportunities for batching or parallel execution. 
Future work should investigate parallel scheduling, so that wall-clock time can be reduced without sacrificing the current instance segmentation logic.

\section{Conclusion}
\label{sec:conclusion}

We introduced ReplaySAM, a training-free framework for 3D instance segmentation of particle packs in XCT volumes.
It combines geometry-based interior-point generation with SAM2 mask propagation over three orthogonal slice sequences, using multi-view fusion to form 3D instances. 
On the evaluated real quartz and iron-ore particle pack volumes, ReplaySAM achieved mean instance Dice scores of 0.942 and 0.933, exceeding both expert-assisted watershed and nnInteractive. 
On the synthetic volume, it improved over nnInteractive but remained below watershed, showing that performance depends on particle size and image characteristics. 
Geometry-based prompting substantially reduces computational cost.
The ablations indicate that complementary views and zoomed inference are important for improving segmentation accuracy. 
These results show that pretrained video-segmentation models can provide a practical training-free alternative or initialisation for XCT particle analysis when task-specific annotations and manual prompting are limited.
The main unresolved limitation is small-particle recovery: even with zooming, small particles remain poorly segmented, and prompt quality depends on foreground binarisation. 
Future work should improve low-resolution particle recovery and foreground estimation, while parallelising inference to make the method more scalable.

\section*{Acknowledgements}
The authors thank Nicolas Francois, Neelima Kandula, Aleese Barron and Benjamin Young for preparing the iron-ore and quartz datasets.
The authors also thank Mark Knackstedt for supporting this research.
The authors acknowledge the computational resources provided by Australia's National Computational Infrastructure (NCI), including the Gadi supercomputer.

\bibliographystyle{unsrt}
\bibliography{ref}
\end{document}